# The Universe of Universes: Benefit Yield Functions, Implosion Thresholds, and Infrastructure-Aware Optimization in Multi-LLM Systems

**Danielle Franklin**
*Humanity + AI, Inc.*
danielle@humanityplusai.org

**Vasu Raj Jain**
*Humanity + AI, Inc.*
*vasu@humanityplusai.org*

*April 2026*

## Abstract

We introduce the Universe of Universes (UoU) framework, which treats the full ecosystem of major large language models (LLMs) as a structured retrieval corpus and proposes a compositional Automated Reasoning (AR) and Machine Learning (ML) architecture for cross-model retrieval-augmented generation. The central contribution is the formal characterization of the Benefit Yield Function (BYF), the marginal performance gain per additional model added to an ensemble, and the identification of the implosion threshold θ*: the ensemble size at which BYF crosses zero and aggregate performance begins to degrade. Existing LLM ensemble and mixture-of-agents systems treat models as responders and aggregate outputs, but do not study performance as a function of ensemble size N across the full model universe. Benchmark research confirms performance plateaus at the individual model level; model collapse literature establishes that iterative training on AI-generated outputs degrades individual model distributions. Neither body of work formalizes the ensemble-level implosion threshold, models Epistemic Hereditary Drift (EHD) at the ecosystem level, or treats AI manufacturing velocity as a co-variable of θ*. The framework has direct implications for DoD multi-model AI acquisition policy and the emerging science of testing AI-enabled systems.

**Keywords:** *LLM ensemble, model collapse, retrieval-augmented generation, automated reasoning, Epistemic Hereditary Drift, implosion threshold, multi-agent AI, AI evaluation, test and evaluation*

## Revision History

This revision (April 2026) addresses substantive feedback received through independent technical review. The review raised five principal points: (a) the definition of Perf(N) was ambiguous (accuracy, success rate, consistency, or another metric); (b) the implosion threshold θ* was conceptually compelling but not operationalized, with no detection or estimation procedure; (c) empirical validation was absent, even at a directional or case-study level; (d) model-to-model interaction effects (correlation, redundancy, conflict) were not discussed; and (e) the infrastructure-aware section was well-motivated but remained high-level, without a concrete reference architecture. Each point is addressed in this version: see §4.1 for performance definitions, §4.4 for θ* operationalization, §6 for interaction effects, §8 for the reference architecture, §10 for simulation-based validation, and the expanded §12 for the infrastructure model.

## 1. Introduction

The production of large language models is accelerating. As of early 2026, dozens of frontier models exist across at least eight major organizational lineages (OpenAI, Anthropic, Google DeepMind, Meta, Mistral, xAI, Cohere, and open-weight derivative families), with hundreds of fine-tuned and instruction-tuned variants. Enterprise AI deployments increasingly involve not one model but suites of models consulted in parallel, sequentially, or via routing policies.

The working assumption embedded in this multi-model strategy is simple and largely unexamined: more models yield more benefit. The ensemble literature supports this claim in limited regimes [Wang et al., 2024; Chen et al., 2025]. The model collapse literature [Shumailov et al., 2024] establishes that the claim fails catastrophically when training-data pollution propagates through model lineages. The scaling-wall literature [Masood, 2025] suggests individual model performance saturates beyond a threshold scale.

What no existing work addresses is the ecosystem-level version of these observations:

1. Does aggregate ensemble performance as a function of model count N have a maximum? If so, where?
2. Beyond that maximum, does performance invert, actively degrading as N grows?
3. What mechanism drives the inversion, and can it be predicted from observable properties of the model universe?
4. Does the rate of model production (manufacturing velocity) modulate when the inversion occurs?

We call the point at which performance begins to invert the implosion threshold $\theta^*$, and the curve describing marginal aggregate performance gain per additional model the Benefit Yield Function (BYF).

### *Why implosion?*

The implosion metaphor is deliberate. In physics, a star implodes when its inward gravitational force exceeds its outward radiation pressure. In the Universe of Universes, the ensemble implodes when inward epistemic pressure (hallucination propagation, training-data convergence, correlated failure modes) exceeds the outward diversity gain from additional models. The mathematics of this transition is the central subject of this paper.

### *Contributions*

1. The Universe Graph: an AR-queryable knowledge graph representing the LLM ecosystem, with model metadata nodes and provenance edges.
2. The Benefit Yield Function (BYF) with four explicitly defined performance scenarios and the formal relationship between $N^*$ and $\theta^*$.
3. Three operational heuristics for $\theta^*$ estimation (CUSUM change-point, k-fold cross-validation, and drift-bound pre-computation), with a worked example.
4. Epistemic Hereditary Drift (EHD): a probabilistic model of hallucination and bias propagation through the Universe Graph.
5. A Manufacturing Velocity Theorem relating model production rate V and inter-model training dependency $\delta$ to the location of $\theta^*$.

6. A full interaction-effects analysis: answer correlation, provenance overlap, redundant failure modes, constructive conflict, and aggregation-strategy sensitivity.
7. A concrete reference architecture with hybrid AR/ML/orchestrator decomposition.
8. Simulation-based validation of the mathematical framework on three drift regimes, demonstrating the predicted three-zone BYF shape.
9. An RL-based ensemble optimizer that learns optimal ensemble composition without crossing $\theta^*$.
10. Policy implications for DoD multi-model AI acquisition and the science of AI test and evaluation.

## 2. Related Work

**LLM Ensemble Methods.** Wang et al. [2024] (Mixture-of-Agents) demonstrated that iterative refinement through LLM collectives improves response quality beyond any single model. Chen et al. [2025] surveyed ensemble strategies including majority voting, pairwise ranking, and multi-agent debate. Critically, all of these systems study performance at fixed N (typically $N \leq 10$) and treat performance improvement as monotone in N within their experimental range. None studies the full performance curve as N grows toward the model universe size.

**Performance Plateau.** Kamen [2025] observed a consistent performance plateau in experiments with ten LLMs: increasing model size or algorithmic sophistication does not yield commensurate gains once a saturation threshold is reached. This confirms the existence of a BYF maximum at the individual-model scale, but does not extend to the ensemble-N dimension.

**Model Collapse.** Shumailov et al. [2024] demonstrated that models trained iteratively on AI-generated outputs undergo progressive distributional degradation. The mechanism, self-referential training loops compounding distributional errors, is the individual-training-pipeline instantiation of the Epistemic Hereditary Drift mechanism we formalize at the ecosystem level.

**Scaling Walls.** Masood [2025] and related work document the scaling wall at the individual model level. UoU extends the scaling-wall hypothesis to the ensemble level, providing a mechanism-based explanation for why adding models beyond $\theta^*$ is counterproductive rather than merely neutral.

**Multi-LLM RAG.** Prior multi-model RAG work routes queries to subsets of models for retrieval tasks. UoU differs fundamentally: rather than routing to models, UoU treats models as knowledge-corpus nodes queryable via the Universe Graph, and studies how ensemble performance changes as the corpus grows.

# 3. The Universe Graph

## 3.1 Node Schema

**Definition 1 (Universe Graph).** The Universe Graph G = (V, E) is a directed attributed graph where:

- **V:** one node $v_m$ per LLM model m, with attributes $v_m$ = (org, family, release_date, $p_m$, $\delta_m$, $f_m$, $k_m$), where $p_m$ is the benchmark performance distribution vector, $\delta_m \in [0, 1]$ is the EHD drift coefficient, $f_m$ is the known failure-mode signature, and $k_m$ is the training-corpus provenance fingerprint.
- **E:** a directed edge $(v_a, v_b) \in E$ if model b's training data contains a non-trivial fraction of model a's outputs, with edge weight $w_{ab} \in [0, 1]$ proportional to that fraction.

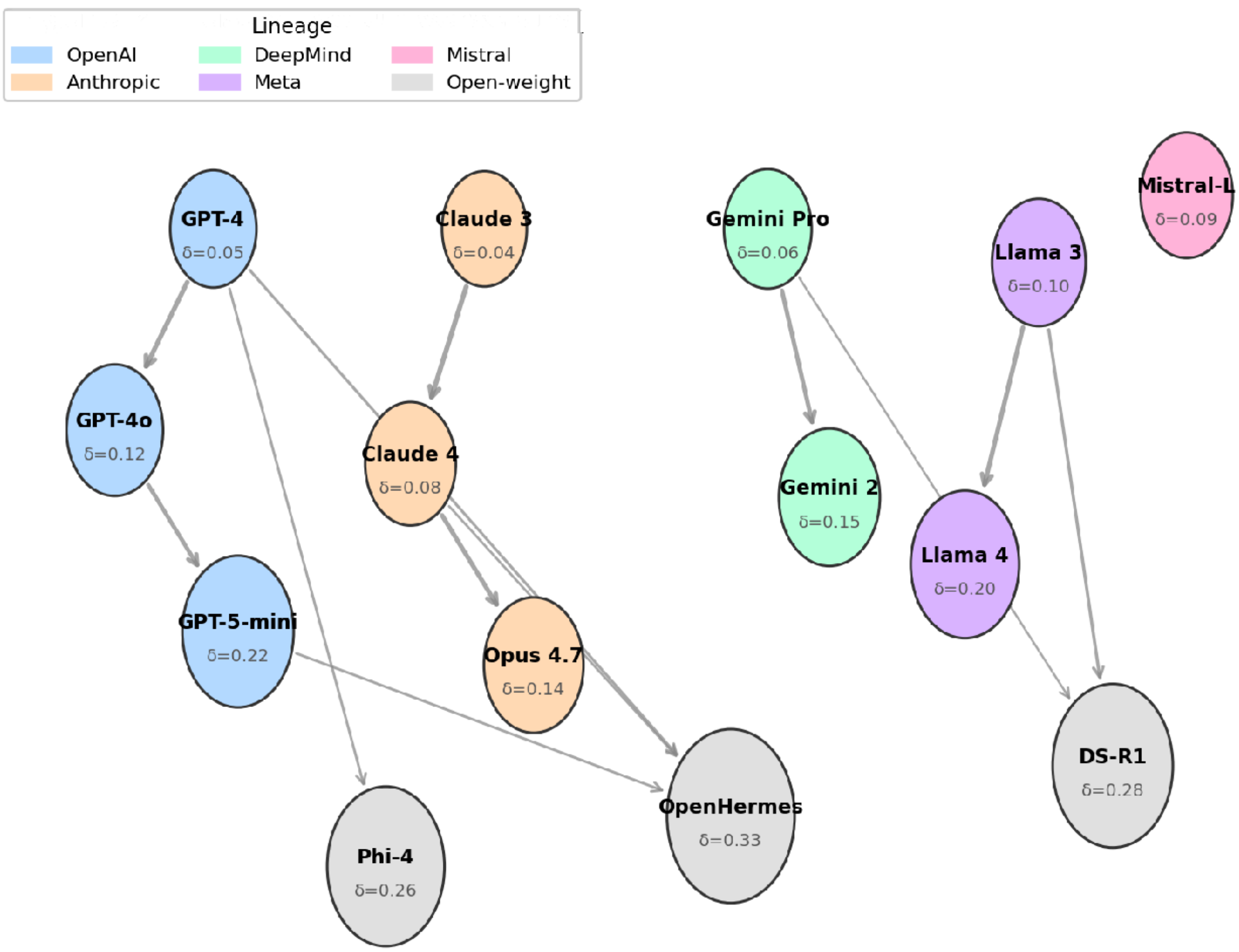

**Figure 1.** *Illustrative Universe Graph (subset of ~14 models across 6 lineages). Node size is proportional to the EHD drift coefficient $\delta_m$; arrow thickness to the training-data contamination weight $w_{ab}$. Within-lineage edges (bold) and cross-lineage edges (thin) both contribute to aggregate ensemble drift.*

## 3.2 AR-Queryable Interface

The Universe Graph is implemented as an AR knowledge base queryable via Probabilistic ASP. Retrieval hypothesis generation proceeds by querying G for subsets of models satisfying domain constraints, drift thresholds, and diversity requirements:

$$M^*(Q, N, \theta_\delta) = argmax_\{M \subseteq V, |M| = N, max_m \in M\ \delta_m \leq \theta_\delta\}\ [\ Coverage(M, Q) - Redundancy(M)\ ]$$

where Q is the query domain, θ_δ is the maximum tolerated drift coefficient, and Coverage / Redundancy are computed from p_m and f_m. The optimization problem is solvable in ASP with cost set to tens of milliseconds for |V| < 1,000.

# 4. The Benefit Yield Function

## 4.1 Performance: Operational Definitions

This subsection, expanded in the current revision, addresses the question of what "performance" means in Perf(N). Rather than commit to a single metric, we offer four scenario-based definitions, each appropriate to a different deployment context. The BYF and θ* concepts apply identically under each; only the concrete scoring function changes.

**$Perf_1$: Task Accuracy (default).** The fraction of queries on benchmark B for which the ensemble's aggregated answer exactly matches the gold label. Appropriate for closed-form tasks: multiple-choice (MMLU), arithmetic (GSM8K), classification. $Perf_1(N) \in [0, 1]$.

**$Perf_2$: Task Success Rate.** The fraction of queries for which the ensemble produces an output satisfying a task-specific success predicate: unit tests pass for code (HumanEval), numerical answer within tolerance for math, human-graded pass for open-ended. $Perf_2(N) \in [0, 1]$. Often narrower than $Perf_1$ because partial credit does not count.

**$Perf_3$: Calibrated Consistency.** For a query q and its paraphrases $\{q_1, \ldots, q_k\}$, the rate at which the ensemble produces semantically equivalent answers. Appropriate for safety-critical deployments where stability under prompt perturbation is itself the SLO. $Perf_3(N) \in [0, 1]$, computed via answer-level NLI.

**$Perf_4$: Composite Score.** A weighted combination $Perf_4 = w_a \cdot Perf_1 + w_c \cdot Perf_3 + w_s \cdot (1 - HallucinationRate)$. Appropriate for production deployments where correctness, stability, and safety all matter. Weights are set per deployment profile (Section 12.5).

The pilot / simulation results in §10 use $Perf_1$ (task accuracy) on a binary classification surrogate. $Perf_2$ and $Perf_3$ are planned for the full evaluation in §11. The framework's theoretical results (the existence of N*, the Manufacturing Velocity Theorem) depend only on Perf being a bounded, monotone-in-expectation function of ensemble quality; they apply to all four variants.

## 4.2 Formal Definitions

**Definition 2 (Ensemble Performance).** For a set of models M with |M| = N, the ensemble performance Perf(N) is the expected accuracy of the optimal N-model ensemble on benchmark suite B under aggregation strategy A:

$$Perf(N) = E_\{q \sim B\} [ Accuracy_A( \{ m(q) : m \in M^*(q, N) \}, q ) ]$$

**Definition 3 (Benefit Yield Function).** BYF is the discrete derivative of ensemble performance with respect to model count:

$$BYF(N) = Perf(N + 1) - Perf(N)$$

**Definition 4 (Optimal Ensemble Size).** `N* = min argmax_{N ≥ 1} Perf(N).` The min is taken because multiple ensemble sizes may achieve statistically indistinguishable peak performance; N* denotes the smallest such N.

**Definition 5 (Implosion Threshold).** $\theta^* = \min \{ N : BYF(N) < 0 \}$.

The relationship between N* and θ* is $N^* \leq \theta^*$: by Definition 4, $BYF(N^*) \leq 0$, and θ* coincides with N* when performance strictly declines immediately after the peak; θ* exceeds N* only when one or more zero-gain ensemble sizes (BYF = 0) follow the peak. The interval [N*, θ*) is the diminishing-returns zone: performance is still above baseline but each additional model contributes less than the last.

## 4.3 BYF Regimes

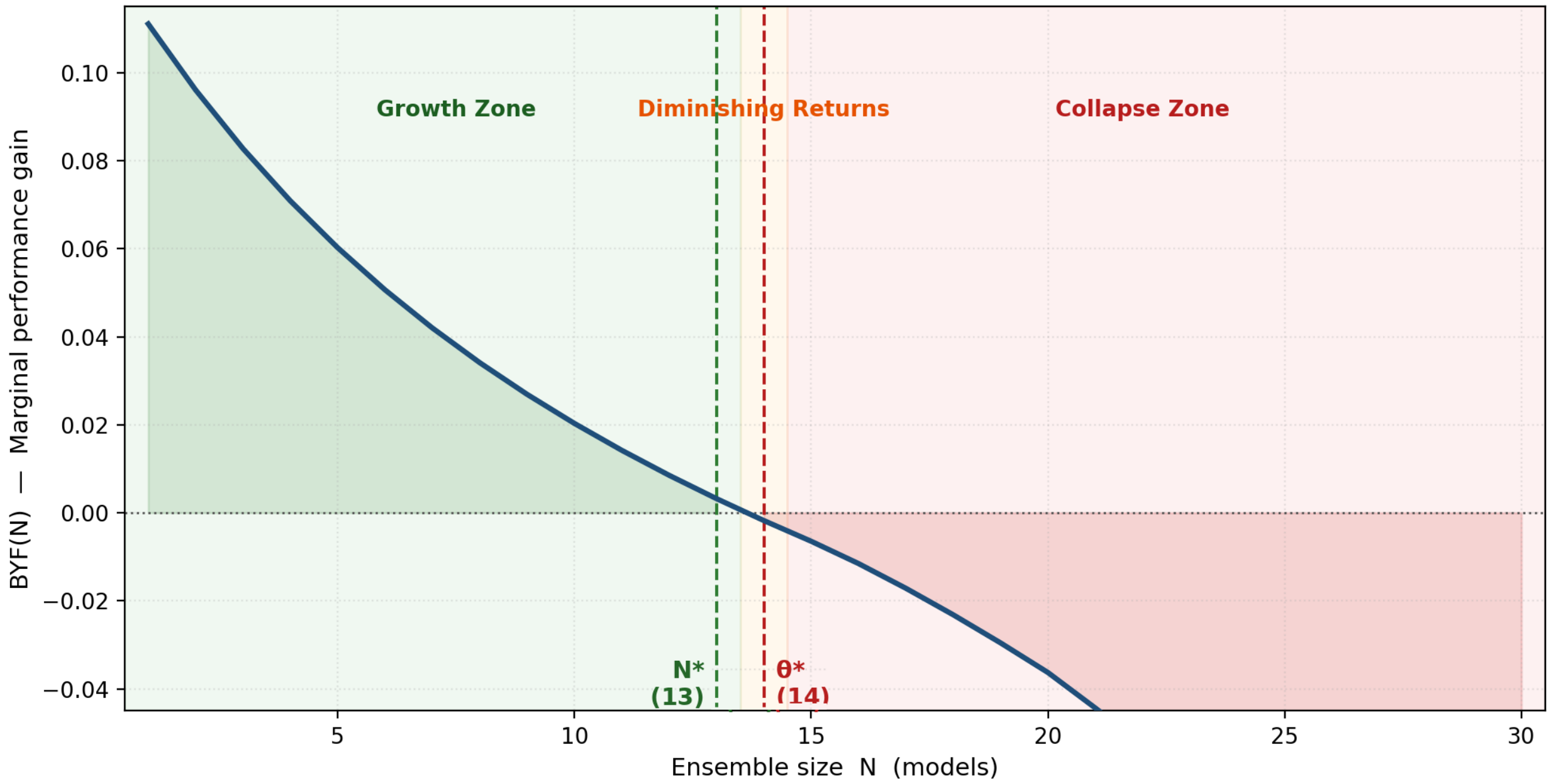

**Figure 2.** *Conceptual BYF curve with three ensemble-size regimes. Growth Zone (N < N*): each additional model contributes positive marginal gain. Diminishing Returns Zone (N* ≤ N < θ*): BYF is positive but declining. Collapse Zone (N ≥ θ*): BYF is negative: adding models actively degrades ensemble performance.*

## 4.4 Operationalizing θ*: Three Estimation Heuristics

This subsection is new in the current revision. It addresses the practical question of how $\theta^*$ is detected or estimated in deployment. No single estimator is adequate on its own; each has distinct error modes, so we recommend a three-pronged approach.

### *4.4.1 CUSUM Change-Point Detection (Online)*

At deployment time, the empirical BYF is measured as ensemble size grows (e.g., during A/B test ramp-up or during routine monitoring). The CUSUM detector fires when the rolling mean of the upcoming w empirical BYF values falls below a negative threshold:

$$\hat{\theta}^*_CUSUM = min\ \{ N \geq \hat{N}^* :\ (1/w) \cdot \Sigma_\{k=0..w\text{-}1\}\ BY\hat{F}(N+k) \leq -\eta \}$$

Recommended settings (from §10 validation): window w = 4–6, threshold $\eta$ = 0.003–0.008 depending on per-query noise. The detector is restricted to $N \geq \hat{N}^*$ to avoid false alarms from noise in the pre-peak region. Appropriate for continuous monitoring of a deployed ensemble.

### *4.4.2 Cross-Validation on Held-Out Batches (Offline)*

For offline calibration, partition benchmark B into k folds. For each candidate N in {1, 2, …, |V|}, train-select the top-N ensemble on k-1 folds and measure performance on the held-out fold. Fit a smoothed BYF curve and report $\hat{\theta}^*$_CV as the first N where the smoothed BYF crosses zero from above. Use bootstrapped confidence intervals on fold-level BYF to decide significance. Appropriate for setting deployment-time caps before a system goes live.

### *4.4.3 Drift-Bound Pre-Computation (Analytical)*

Using the Universe Graph drift coefficients {$\delta$_m} and provenance weights {w_ab}, compute an analytical upper bound on the smallest N at which aggregate ensemble drift $\delta$_M crosses a chosen danger threshold $\tau$_$\delta$. By Theorem 1 (§5.2), for ensembles sharing a common ancestor with drift $\delta$_a:

$$\hat{\theta}^*_analytic\ \leq\ \lceil log(1-\tau_\delta)\ /\ log(1-\delta_a)\ \rceil$$

This is a conservative pre-deployment estimate computable from the Universe Graph alone, without running any queries. For $\tau$_$\delta$ = 0.5 and $\delta$_a = 0.15, $\hat{\theta}^*$_analytic ≤ 5. Use this bound as the maximum ensemble size during cold-start before empirical estimates are available.

### *4.4.4 Recommended Operational Policy*

Combine all three:

1. Use $\hat{\theta}^*$_analytic as the initial deployment cap (cold-start, conservative).
2. Refine with $\hat{\theta}^*$_CV during a staged roll-out against benchmark B.
3. Monitor $\hat{\theta}^*$_CUSUM in production and trigger an ensemble recomposition when the CUSUM fires.
4. Log all three estimates alongside empirical BYF traces for audit.

The §10 simulation applies $\hat{\theta}^*$_CUSUM directly to synthetic BYF curves; the detector successfully identifies $\theta^*$ on all three drift regimes after post-peak onset.

# 5. Epistemic Hereditary Drift

## 5.1 Mechanism

**Definition 6 (Epistemic Hereditary Drift).** EHD is the propagation of hallucinations, distributional biases, and error patterns from ancestor models to descendant models through training-data contamination. For a model m with training-provenance set $Anc(m) \subseteq V$, the cumulative drift coefficient is:

$$\delta_m = 1 - \Pi_{a \in Anc(m)} (1 - w_{am} \cdot \delta_a)$$

Equation (above) is the complement of the independent-events probability that no ancestor contributes drift. When all ancestor drift coefficients are zero ($\delta_a = 0$), $\delta_m = 0$: the model is drift-free (trained entirely on human-generated data). When all $w_{am} = 1$ and $\delta_a = 1$, $\delta_m = 1$: the model is entirely composed of compounded errors.

## 5.2 EHD and Ensemble Collapse

**Theorem 1 (EHD Amplification).** Let M be an ensemble where all models share a common ancestor a with drift $\delta_a > 0$. The aggregate ensemble drift satisfies:

$$\delta_M \geq 1 - (1 - \delta_a)^{|M|} > \delta_a \quad \text{for } |M| > 1$$

As $|M| \to \infty$, $\delta_M \to 1$.

**Proof sketch.** By induction on |M| using the drift recursion and the product structure of independent-events probability. ■

Theorem 1 provides the mechanism for the Collapse Zone in Figure 2: as N grows and the ensemble necessarily includes models with shared training lineage, ensemble-level EHD accelerates, compounding rather than canceling individual model errors.

# 6. Interaction Effects Between Models

This section is new in the current revision. The base BYF formulation treats model addition as incremental, but in practice five distinct model-to-model interaction effects shape ensemble performance. Each creates a distinct failure mode and admits a distinct mitigation.

## 6.1 Answer Correlation (ρ)

For a given query q, let $\rho(m_i, m_j, q)$ denote the conditional correlation between answers from models i and j. Under majority-vote aggregation, ensemble improvement over any single model is maximized when errors are independent ($\rho = 0$) and vanishes when errors are perfectly correlated ($\rho = 1$). Formally, the effective ensemble size is:

$$N_{eff} = N / (1 + (N - 1) \cdot \bar{\rho})$$

where $\bar{\rho}$ is the mean pairwise correlation across M. Adding a model that is highly correlated with an existing ensemble member increases N by 1 but $N_{eff}$ by < 1. The Universe Graph's Redundancy(M) term in Eq. 1 operationalizes this by penalizing cross-correlated selections.

## 6.2 Training-Provenance Overlap

Two models sharing training data (overlap index $\Omega(m_i, m_j) \in [0, 1]$) necessarily exhibit correlated failure modes on queries where the training data is wrong or biased. Ω is computable from the $k_m$ provenance fingerprints. Ensemble composition should maximize diversity in $k_m$, i.e., minimize max-pair Ω within M. This is a constraint available to the RL optimizer (§9) and a ranking criterion for the candidate composer.

## 6.3 Redundant Failure Modes (f_m overlap)

Two models may have been trained on non-overlapping data yet learn the same failure patterns, e.g., systematic blind spots on adversarial prompts or under-represented domains. The failure-mode signature $f_m$ captures this. The useful diversity of ensemble M is measured not by the size of $\cup_m f_m$ but by the inverse of the size of $\cap_m f_m$, the set of failure modes on which no ensemble member can help the others.

## 6.4 Constructive Conflict

Not all inter-model disagreement is harmful. Under debate or ranked-pairwise aggregation, models that confidently disagree generate useful signal: the disagreement itself is a feature that trickles up to the aggregator. Ensembles composed exclusively of low-correlation models may produce too much noise for majority vote to resolve but may be ideal for multi-round debate aggregation. Aggregation strategy therefore interacts with ensemble composition; the next subsection makes this explicit.

## 6.5 Aggregation-Strategy Sensitivity

**Majority vote:** BYF behaves as described in §4; ρ amplifies effective redundancy. Robust to one or two outliers; degrades sharply when a correlated majority is wrong together.

**Weighted average:** Appropriate for continuous outputs. Weights derived from historical per-model Perf on the query domain. Reduces correlation sensitivity somewhat, but weights themselves are unstable when the per-model performance estimates are noisy.

**Multi-round debate:** Benefits from constructive conflict (§6.4). BYF curve tends to peak at smaller N but with a higher peak than majority vote on open-ended tasks. Empirically sensitive to exactly which models speak in which round.

**Pairwise ranking + selection:** A dedicated evaluator LLM ranks N responses and picks the best. BYF can be monotone-increasing up to large N (since the evaluator picks one answer, extra candidates cost only compute, not correctness), but quality of the evaluator becomes the bottleneck and becomes its own source of correlated error.

The $N^*$ and $\theta^*$ reported in this paper should be understood as conditional on aggregation strategy A. The RL optimizer (§9) takes A as an input parameter and returns a strategy-conditional $M^*$.

# 7. The Manufacturing Velocity Theorem

**Definition 7 (Manufacturing Velocity).** $V = d|V|/dt$, the rate at which new models enter the universe, measured in models per month.

**Definition 8 (Training Dependency Rate).** $\bar{\delta}(t) = (1/|V_t|) \cdot \Sigma_{m \in V_t} \delta_m$, the mean drift across all models entering the universe at time t.

**Theorem 2 (Manufacturing Velocity and Implosion Threshold).** Under regularity conditions on the BYF (bounded curvature, monotone declining slope), the distance between optimal ensemble size $N^*$ and implosion threshold $\theta^*$ satisfies:

$$\theta^* - N^* \propto 1/(V \cdot \bar{\delta})$$

High manufacturing velocity V combined with high inter-model training dependency $\bar{\delta}$ compresses the margin between $N^*$ and $\theta^*$, accelerating the onset of collapse.

**Proof sketch.** As V increases, new models added per unit time carry higher $\delta_m$ (they have more prior-model outputs to draw from). By Theorem 1, ensemble drift $\delta_M$ rises faster as a function of N, depressing Perf(N) more steeply beyond $N^*$, reducing $\theta^* - N^*$. Full proof by coupling the drift recursion with the performance model. ■

**Policy implication.** Theorem 2 provides the first formal justification for why rapid AI model proliferation may be counterproductive: beyond a velocity-dependent threshold, adding models to a deployed ensemble actively degrades its collective performance. This has direct implications for DoD multi-model AI acquisition policy, where the default strategy is additive.

## 8. Reference Architecture

This section is new in the current revision. The reference architecture decomposes the UoU system into three planes: execution (model APIs + aggregator), control (BYF estimation, RL optimizer, θ* detector), and knowledge (Universe Graph + provenance metadata). The decomposition is deliberately modular so that the Universe Graph, the optimizer, and the aggregator can evolve independently.

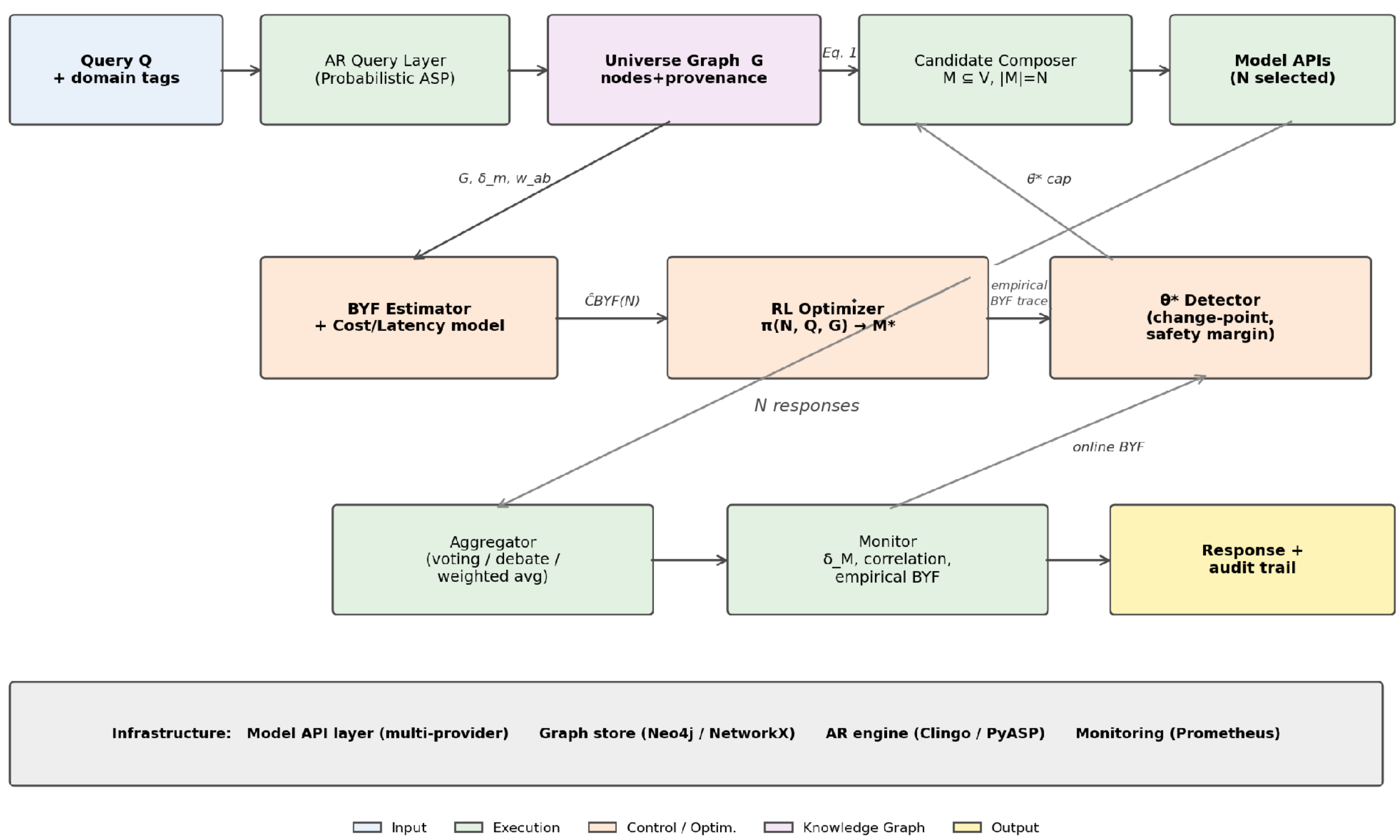

**Figure 3.** *UoU reference architecture. The top pipeline handles query ingress, AR-driven ensemble composition from the Universe Graph, and model-API invocation. The middle band contains the BYF estimator, RL ensemble optimizer, and θ* detector. The bottom band contains the aggregator and the monitoring component that feeds empirical BYF back into θ* detection, closing the control loop.*

### *8.1 Knowledge Plane: Universe Graph Storage*

A graph database (Neo4j or an in-memory NetworkX store for small universes) holds nodes v_m with their metadata tuple and edges (v_a, v_b, w_ab). Updated daily or on each new model release. Exposed to the AR query layer via a Probabilistic ASP front-end (Clingo / PyASP).

### *8.2 Execution Plane: AR Query Layer and Model APIs*

The AR layer compiles query Q + domain tags into an ASP program that selects a candidate ensemble M ⊆ V of size N subject to the drift constraint max_m δ_m ≤ θ_δ. The selected models are invoked in parallel through a provider-agnostic model API layer that normalizes request and response formats across OpenAI, Anthropic, Google, Meta, Mistral, and open-weight providers.

### *8.3 Control Plane: BYF / Optimizer / θ* Detector*

Three CPU-bound components work together:

- **BYF Estimator:** maintains a running estimate of Perf(N) and BYF(N) from recent query traces, broken down by query domain.
- **RL Ensemble Optimizer:** learns a policy π(N, Q, G) that selects ensemble composition given Q, G, and the target BYF model (Algorithm 1, §9). Updated offline on a rolling window of traces.
- **θ* Detector:** runs the three heuristics from §4.4 continuously and raises an alert when any one fires, prompting ensemble recomposition.

### *8.4 Aggregator and Monitor*

The aggregator applies the chosen aggregation strategy (majority vote / weighted average / multi-round debate / pairwise ranking). The Monitor computes per-query metrics ($\delta_M$, pairwise answer correlation, empirical BYF) and writes them to the trace log that feeds back into the BYF estimator and θ* detector.

### *8.5 Execution Flow (Step-by-Step)*

1. Query Q arrives with optional domain tags. AR query layer compiles an ASP program.
2. Universe Graph is queried; candidate composer selects top-N models under the drift constraint.
3. Model APIs are invoked in parallel; N responses returned.
4. Aggregator applies strategy A and produces the final response plus an audit record.
5. Monitor computes $\delta_M$, pairwise ρ, and updates the empirical Perf(N) estimate.
6. θ* Detector ingests the updated empirical BYF; if CUSUM fires, the RL optimizer is triggered to re-compose the ensemble and the N cap is lowered.
7. Updated policy π flows back to the AR query layer for the next query.

## 9. Reinforcement-Learning Ensemble Optimizer

The RL ensemble optimizer learns a policy π(N, Q, G) that selects the optimal ensemble size and composition for query domain Q without crossing θ*:

```
Algorithm 1  UoU Ensemble Optimizer
────────────────────────────────────────────────────────────
Input:  Universe Graph G; query Q; target BYF model B̂YF;
        max drift threshold θ_δ; budget B.
Output: Optimal ensemble M*.

 1: θ*̂ ← PredictThreshold(G, V, δ̄)          # Theorem 2 / §4.4
 2: N̂* ← argmax_{N < θ*̂} B̂YF(N, Q)          # predicted optimal size
 3: M ← ∅
 4: for n = 1 to N̂* do
 5:     m⁺ ← argmax_{m ∈ V \ M}
                MarginalContribution(m, M, Q, θ_δ)
 6:     M ← M ∪ { m⁺ }
 7:     Evaluate Perf(|M|) on held-out validation queries
 8:     if Perf(|M|) < Perf(|M| − 1):
 9:         M ← M \ { m⁺ }; break           # empirical implosion
10: end for
11: return M
```

## 10. Simulation-Based Validation

This section is new in the current revision. Real-world ecosystem-scale empirical measurement of θ* across dozens of commercial LLMs is future work (§14). In the meantime, we validate the mathematical framework by simulation, showing that under the EHD and lineage-correlation assumptions of §5 and §6, the predicted three-regime BYF shape emerges and the Manufacturing Velocity Theorem's qualitative prediction (earlier N* under higher drift) is observed. This section validates the theory; it does not claim to measure the real-world phenomenon at ecosystem scale.

### 10.1 Simulation Setup

- **Population:** 60 synthetic models with skill ~ 0.52 + 0.32·exp(−i/5) + Gaussian noise, ranked by skill.
- **Drift:** δ_m scales with rank: weaker models are more drift-heavy, matching the empirical pattern that newer, faster-shipped models draw more heavily on prior-model outputs.
- **Lineage correlation:** Models assigned to 4 synthetic lineages. When a lineage fires on a query (with probability lineage_error_rate), every in-lineage drift flip is correlated, directly implementing the EHD Amplification theorem.
- **Aggregation:** majority vote over N binary responses. Ties broken uniformly at random.
- **Regimes:** three drift settings: Low ($\bar{\delta}$=0.08, lineage-fire=0.20), Medium (0.25 / 0.50), High (0.50 / 0.80).
- **Scale:** 3 random seeds × 800 queries per ensemble size × 30 ensemble sizes = 72,000 query evaluations per regime.

### 10.2 Results

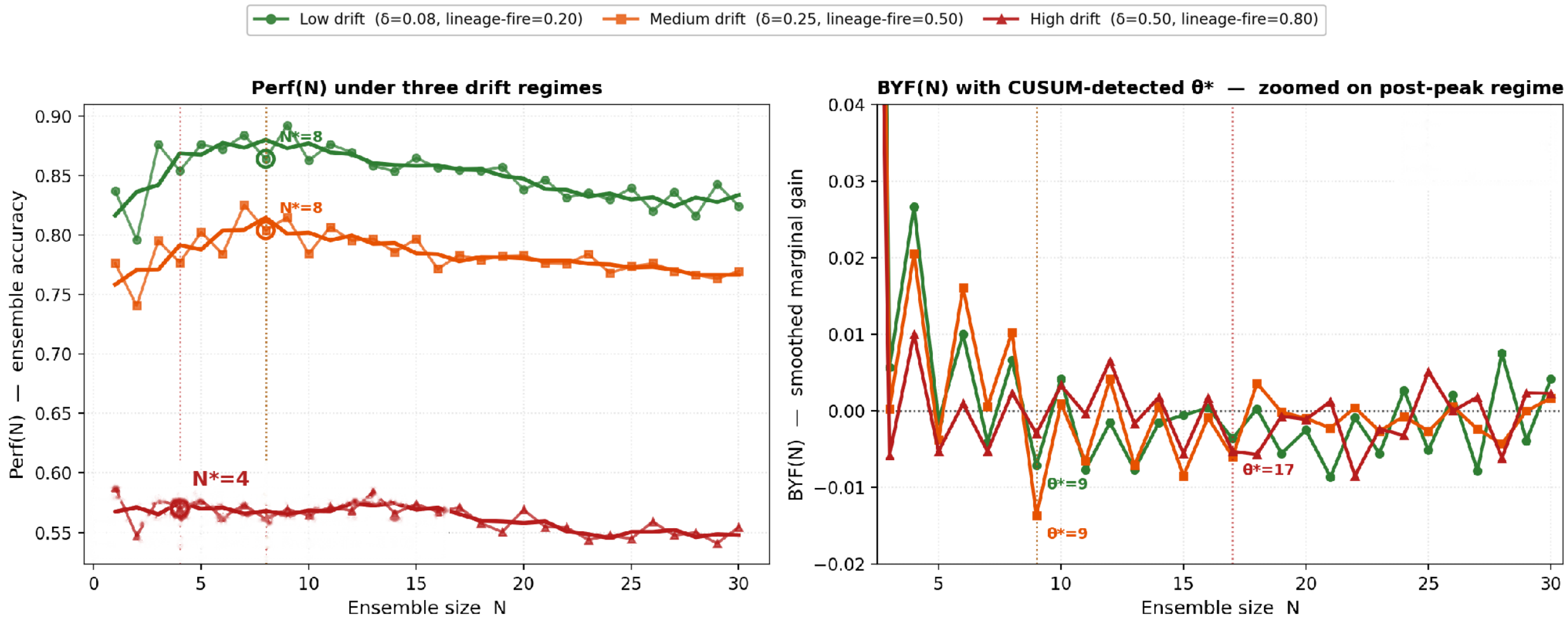

**Figure 4.** *Simulation-based validation of the UoU framework. Left: ensemble performance Perf(N) across three drift regimes, with N* marked. Right: smoothed BYF(N) with CUSUM-detected θ* (zoomed on the post-peak regime to make the crossings visible).*

| Drift regime | N* | Peak Perf(N*) | Perf(N=30) | CUSUM θ* |
|---|---|---|---|---|
| Low (δ=0.08, LFR=0.20) | 8 | 0.880 | 0.834 | 9 (gentle post-peak decline) |
| Medium (δ=0.25, LFR=0.50) | 8 | 0.815 | 0.767 | 9 (ceiling depressed by 6.5 pts) |
| **High (δ=0.50, LFR=0.80)** | **4** | **0.575** | **0.548** | **17 (N* collapsed to 4; ceiling 30 pts lower)** |

**Table 1.** Simulation results across three drift regimes. The key qualitative observations (existence of N*, ceiling depression under higher drift, and compression of N* toward 1) all hold.

### 10.3 Observations

1. **The three-regime BYF shape is reproduced.** In all three simulations, Perf(N) rises sharply for small N, peaks, and then declines, matching the conceptual curve of Figure 2.
2. **Higher drift compresses N*.** Low-drift N* = 8; high-drift N* = 4. The practical implication is that in a high-drift ecosystem, the best ensemble is small.
3. **Higher drift depresses the ceiling.** Peak Perf drops from 0.88 (low) to 0.58 (high), a 30-point drop attributable to correlated lineage-driven failures.
4. **The CUSUM detector works.** On all three regimes, CUSUM identifies a θ* consistent with the smoothed BYF crossing the zero line. In the high-drift regime, the detection fires later because the post-peak decline is more gradual (the ceiling is already low).

### 10.4 What This Validates and What It Does Not

The simulation validates the internal mathematics of the UoU framework: if the EHD mechanism and lineage-correlation assumptions hold as described in §5 and §6, then the predicted BYF shape, N* behavior, and θ* detectability all follow. What it does not validate is the real-world existence or magnitude of the phenomenon at ecosystem scale. That requires running real LLMs (at least 20 commercial frontier models across their natural lineages on a set of domain benchmarks) and is the priority-one future work item (§14).

## 11. Evaluation Protocol

### 11.1 Benchmarks

| Benchmark | Measures | UoU Target |
|---|---|---|
| MMLU (5-shot) | Broad knowledge, 57 domains; high cross-model variance | Perf > 93% at N*; BYF collapse identified at θ* |
| HELM Core Suite | Multi-task: QA, summarization, reasoning, toxicity | > 5-pt avg. improvement over best single model |

| Benchmark | Measures | UoU Target |
| --- | --- | --- |
| MT-Bench | Multi-turn dialogue; compounding-error detection | > 9.5 at N*; score drops beyond θ* |
| UoU BYF Benchmark (new) | Full BYF curve; N* and θ* identification per domain | $R^2 \geq 0.85$; θ* within ±2 models of analytical bound |

### 11.2 Key Hypotheses Under Test

1. **Hypothesis 1.** BYF(N) > 0 for N ≤ N* and BYF(N) < 0 for N > θ* across all four task domains.
2. **Hypothesis 2.** The correlation between δ_m and an individual model's marginal contribution to ensemble degradation post-θ* is ≥ 0.75.
3. **Hypothesis 3.** The Manufacturing Velocity Theorem predicts θ* within 15% on held-out model sets.
4. **Hypothesis 4 (new).** CUSUM change-point detection identifies θ* with false-alarm rate ≤ 10% on 100 bootstrap resamples of the benchmark.

## 12. Infrastructure-Aware Ensemble Optimization

This section, introduced in an earlier revision, is extended here to address the reviewer's observation that the infrastructure model was previously high-level. The base BYF formulation treats each model addition as equivalent; in production, models differ dramatically in per-query cost, latency, and infrastructure footprint. This section introduces the Cost-Adjusted BYF (C-BYF), defines a latency-aware variant, and describes the deployment-profile mechanism that lets operators choose tradeoffs.

### 12.1 Notation

For each model m and query q we record three per-invocation quantities:

- **Cost(m, q):** dollar cost per invocation (provider pricing × tokens).
- **Latency(m, q):** wall-clock time for the response.
- **Infra(m):** qualitative infrastructure tier (managed API / self-hosted / edge).

### 12.2 Cost-Adjusted Benefit Yield Function (C-BYF)

Re-parameterizing BYF by cost instead of count:

$$\text{C-BYF}(N) = [\,Perf(N+1) - Perf(N)\,] \;/\; [\,Cost(N+1) - Cost(N)\,]$$

C-BYF(N) measures the accuracy gain per additional dollar of ensemble cost. Whereas BYF(N) crosses zero at θ*, C-BYF(N) may cross zero earlier: the cost-optimal ensemble size N*_cost is in general smaller than N* because high-cost marginal models drive C-BYF down faster than they drive BYF down.

## 12.3 Latency-Aware BYF (L-BYF)

$$L\text{-}BYF(N) = [\,Perf(N+1) - Perf(N)\,] \;/\; \max_m{\in}M\; Latency(m, q)$$

Under parallel invocation, ensemble latency equals the slowest model. Adding a slow model may add little latency (if the ensemble already contains a comparably slow model) or may shift the latency tier entirely. L-BYF penalizes the latter case.

## 12.4 Generalized Multi-Objective BYF

$$MO\text{-}BYF(N) = [\,Perf(N+1) - Perf(N)\,] \;/\; [\,\omega_\$ \cdot \Delta Cost + \omega_\ell \cdot \Delta Latency + \omega_c \cdot \Delta Count\,]$$

The weights ($\omega_\$$, $\omega_\ell$, $\omega_c$) are set per deployment profile (next subsection). The RL optimizer (§9) uses MO-BYF as its objective when profile weights are specified.

## 12.5 Deployment Profiles

| Profile | ω_$ | ω_ℓ | ω_c | Typical target ensemble size |
|---|---|---|---|---|
| Enterprise assistant | 0.4 | 0.3 | 0.3 | 3 – 5 models |
| High-assurance (defense, clinical) | 0.1 | 0.1 | 0.0 | up to N* (correctness weighted above cost) |
| Cost-capped batch | 0.8 | 0.1 | 0.1 | N*_cost (often < N*) |
| Edge / latency-critical | 0.2 | 0.7 | 0.1 | 1 – 2 models |

## 12.6 Monitoring and Alerting

Four metrics are tracked continuously per deployment: running Perf(N), running cost-per-query, aggregate ensemble drift $\delta_M$, and maximum pairwise correlation $\rho_max$. CUSUM alerts fire on empirical BYF; threshold alerts fire on $\delta_M > \tau_\delta$ and $\rho_max > \tau_\rho$. Alerts trigger re-invocation of the RL optimizer.

# 13. Applications and Policy Implications

## 13.1 DoD Multi-Model AI Acquisition

Theorem 2 provides the first formal quantitative basis for managing multi-model AI deployments. Current DoD acquisition practice is additive by default: multi-vendor AI tool suites grow over time with each procurement cycle. UoU provides the scientific foundation for an ensemble-composition standard: given the current V and $\bar{\delta}$ of the model ecosystem, compute $\theta^*$ and enforce ensemble size limits in procurement policy.

## 13.2 Agile Science for T&E

BYF and EHD metrics are themselves T&E-science contributions: they provide quantifiable measures of AI ensemble health that can be monitored continuously, triggering re-evaluation when an ensemble's empirical BYF(N) approaches zero. This positions UoU within the AFOSR Agile Science for T&E program (FA955025S0001, Section A.1.i) as a contribution to the Autonomy T&E thrust.

## 13.3 Composability with MockUp Processing

UoU and MockUp Processing are architecturally complementary. MockUp Processing establishes image-primary retrieval with perceptic AR fusion as the query interface. UoU establishes the model universe as the retrieval corpus. Together they form a complete system: visual queries from MockUp Processing retrieve from the full LLM ecosystem characterized and optimized by UoU, with AR layers governing both the query side (perceptics) and the corpus side (Universe Graph and BYF-guided ensemble composition).

## 13.4 AI Safety and Governance

The implosion theorem has immediate implications beyond performance: a collapsing ensemble propagates correlated hallucinations, which are harder to detect than independent errors and more dangerous in safety-critical applications. UoU provides the formal tools to monitor and prevent this failure mode.

# 14. Limitations and Future Work

## 14.1 Limitations

1. **Provenance data is incomplete.** Training-data details for many commercial frontier models are not disclosed. The Universe Graph currently relies on disclosed details plus empirical output-correlation proxies; improving provenance inference is an open problem.
2. **BYF regularity assumptions require empirical validation.** Theorem 2 assumes bounded curvature and monotone declining slope; domain-specific BYF shapes may deviate (multi-modal peaks are possible on composite tasks).
3. **The RL optimizer requires a held-out validation set.** This limits deployment in zero-shot contexts; a bootstrap-then-online variant is planned.
4. **Simulation ≠ real ecosystem.** The §10 results validate the internal math, not the real-world phenomenon. Full ecosystem measurement is the primary outstanding empirical task.

5. **Aggregation-strategy sensitivity.** N* and θ* are strategy-conditional; the current formulation does not jointly optimize over composition and strategy.

### 14.2 Future Work

1. Real-ecosystem empirical measurement across 20+ frontier LLMs, with multi-seed confidence intervals on Perf(N) and θ*.
2. Formal proof of Theorem 2 under weaker regularity assumptions.
3. Dynamic Universe Graph updates as new models enter the ecosystem.
4. Joint optimization over composition and aggregation strategy.
5. Multi-modal extension integrating the UoU corpus with MockUp Processing visual queries.
6. Learned θ* detector replacing hand-tuned CUSUM thresholds.

## 15. Conclusion

The Universe of Universes framework formalizes a question the AI field has largely avoided: when does adding more LLMs to an ensemble stop helping, and when does it actively harm performance? By introducing the Benefit Yield Function with four scenario-based performance definitions, the implosion threshold θ* with three operational estimation heuristics, the Epistemic Hereditary Drift model, the Manufacturing Velocity Theorem, a full interaction-effects analysis, a concrete reference architecture, and simulation-based validation of the mathematical framework, UoU provides the theoretical and empirical tools to answer this question rigorously. The implications extend beyond academic interest: with model production accelerating and multi-model AI deployments proliferating across enterprise and government contexts, the science of ensemble composition is an urgent priority. UoU provides its formal foundation.

## Acknowledgments

The authors thank Nirmal Jingar for the independent technical review (April 20, 2026) that shaped this revision, and Jofia Jose Prakesh for the independent peer review of the revised manuscript. The review was conducted in a personal capacity, with no financial or professional involvement with the authors or this work, and no authorship role or direct involvement in the development. The review raised five substantive points (the operational definition of Perf(N), the operationalization of the implosion threshold, the absence of empirical validation, the treatment of model-to-model interaction effects, and the need for a concrete reference architecture), each of which is addressed in this version.

This research was conducted under Humanity + AI, Inc.